\documentclass[letterpaper, 10 pt, conference]{ieeeconf}  

\IEEEoverridecommandlockouts                         
\usepackage{xcolor}

\usepackage{graphicx}

\usepackage{gensymb}
\usepackage[hyphens]{url}
\usepackage{multirow}
\usepackage{multicol}
\usepackage{tabularx}
\usepackage{dsfont}
\usepackage{url}
\usepackage{color}
\usepackage{subcaption}
\usepackage{caption}
\usepackage{booktabs}
\usepackage{dcolumn}
\usepackage{amssymb}
\usepackage{hyperref}
\usepackage{amsmath}
\usepackage{colortbl}
\usepackage{tikz}
\colorlet{lightgray}{gray!20}

\usepackage{physics}
\usepackage{graphicx}
\usepackage{float}
\usepackage{subfloat}
\usepackage{longtable}
\usepackage{threeparttable}
\usepackage{xcolor}
\usepackage{diagbox}
\usepackage{cite}
\usepackage{graphicx}  
\usepackage{cite}      
\usepackage{xspace}    

\usepackage{color}
\hypersetup{
    colorlinks=false,
    linkcolor=blue,
    filecolor=magenta,      
    urlcolor=cyan,
    pdftitle={Overleaf Example},
    pdfpagemode=FullScreen,
    }

\usepackage{booktabs}
\usepackage{multirow}

 \renewcommand{\paragraph}[1]{
    \vspace{2mm}
     \noindent\textbf{#1} 
 }
\newcommand{\ie}{\textit{i.e.}}

\definecolor{amber(sae/ece)}{rgb}{1.0, 0.49, 0.0}

\title{\LARGE \bf
V2X-DGW: Domain Generalization for Multi-agent Perception under Adverse Weather Conditions}

\author{Baolu Li$^{1}$$^\star$, Jinlong Li$^{1}$$^\star$, Xinyu Liu$^{1}$,  Runsheng Xu$^{2}$, Zhengzhong Tu$^{3}$, Jiacheng Guo$^{1}$, \\
Qin Zou$^{4}$, Xiaopeng Li$^{5}$, Hongkai Yu$^{1}$$^\dagger$ 
\thanks{$^{1}$Cleveland State University. $^{2}$Independent Researcher. $^{3}$Texas A\&M University. $^{4}$Wuhan University. $^{5}$University of Wisconsin-Madison. $^\star$Equal contribution. This research is partially  supported by NSF 2215388 and 2343167. $^\dagger$Corresponding Author: h.yu19@csuohio.edu}  
}

\begin{document}

\maketitle
\thispagestyle{empty}
\pagestyle{empty}

\begin{abstract}
Current LiDAR-based Vehicle-to-Everything (V2X) multi-agent  perception systems have shown the significant success on 3D object detection. While these models perform well in the trained clean weather, they struggle in unseen adverse weather conditions with the domain gap. In this paper,  we propose a Domain Generalization based approach, named \textit{V2X-DGW}, for LiDAR-based 3D object detection on multi-agent perception system under adverse weather conditions. Our research aims to not only maintain favorable multi-agent performance in the clean weather but also promote the performance in the unseen adverse weather conditions by learning only on the clean weather data. To realize the Domain Generalization, we first introduce the Adaptive Weather Augmentation (AWA) to mimic the unseen adverse weather conditions, and then propose two alignments for generalizable representation learning: Trust-region Weather-invariant Alignment (TWA) and Agent-aware Contrastive  Alignment (ACA). To evaluate this research, we add Fog, Rain, Snow conditions on two publicized multi-agent datasets based on physics-based models, resulting in two new datasets: OPV2V-w and V2XSet-w. Extensive experiments demonstrate that our V2X-DGW achieved significant improvements in the unseen adverse weathers.
The code is available at  https://github.com/Baolu1998/V2X-DGW.
\end{abstract}

\section{Introduction}
The advent of multi-agent perception systems has marked a significant leap forward in surmounting the inherent limitations of single-agent perception from the challenges of perceiving range and occlusion~\cite{chen2019f,xu2023v2v4real}. This paper uses the Connected and Automated Vehicles (CAV) based cooperative perception in autonomous driving as an example of the multi-agent perception systems. Leveraging the Vehicle-to-Everything (V2X) communication for the exchange of LiDAR based point cloud data, these multi-agent systems have demonstrated a substantial enhancement in 3D object detection over the single-agent perception systems~\cite{wang2020v2vnet,xu2022opv2v,li2023s2r,li2024advattack,v2xvitv2,gao2025stamp,chenopencda,li2024comamba}. However, the LiDAR-based V2X multi-agent perception systems under the adverse weather conditions (Fog, Rain, Snow)~\cite{zhu2024mwformer} have not been studied and evaluated before.

So far as we know, all the current LiDAR-based V2X multi-agent perception models~\cite{xu2023v2v4real,li2024Break,li2023s2r} are trained on the clean-weather data. In our experiments on the publicized V2X cooperative perception dataset OPV2V~\cite{xu2022opv2v}, if we directly deploy these models (trained on clean weather) to the unseen adverse weather conditions, their performance will drop significantly, for example, AP@IoU=0.5 from 88.10 in clean weather to 67.55 in fog weather. We discover and summarize three specific challenges for the domain gap between clean and adverse weather conditions of multi-agent perception, as detailed below.

\begin{figure}[!t]
\centering
\subfloat{ 
\includegraphics[width=1\columnwidth]{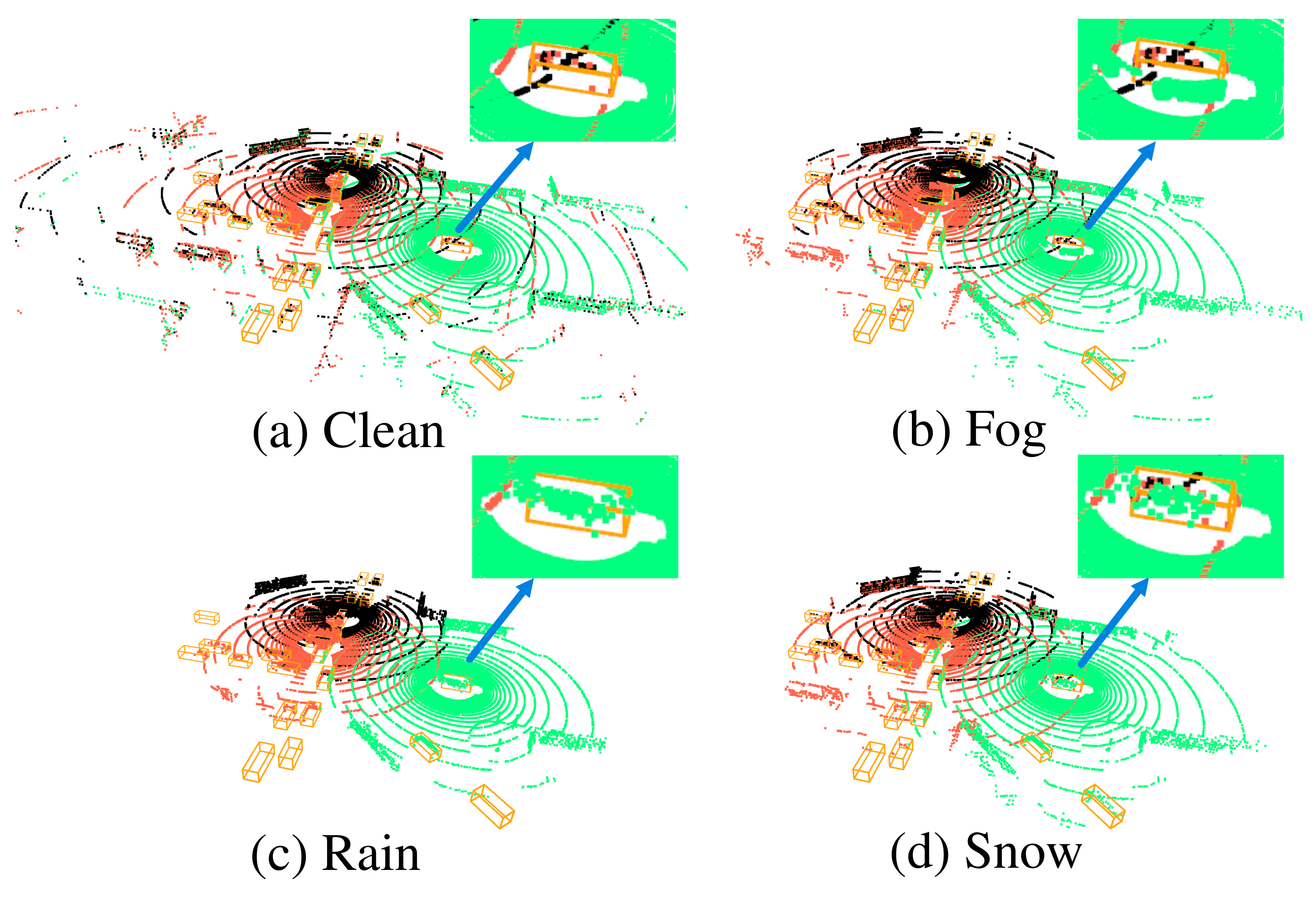}%
}
\caption{\textbf{Multi-agent perception system under adverse weather}. Three CAVs' LiDAR point clouds are highlighted in Black, Orange and Green colors. We can discover the reduced perception range (shrinked wave circles) and degradation (highlighted by the blue arrows) for the point clouds under adverse weather.}
\label{fig:motivation}
\vspace{-1em}
\end{figure}

 \begin{itemize}
    \item \textbf{Perception Range Reduction:} As shown in Fig.~\ref{fig:motivation}, the LiDAR's perception range is reduced in the adverse weather. The fog particles, raindrops, snowflakes occasionally block the LiDAR's line of sight causing lossy data in the point cloud, so this dynamic occlusion reduces the vehicle's ability to perceive its surroundings accurately.

    \item \textbf{Point Cloud Degradation:} As highlighted by the blue arrows in Fig.~\ref{fig:motivation}, the LiDAR's point cloud quality is degraded. Severe weather conditions might change the characteristics compared to clean weather, such as optical attenuation in the atmosphere, surface reflectivity changes, dynamic occlusions. These factors reduce the amount of effective signal returned to the LiDAR, affecting the integrity and accuracy of the point cloud.

    \item \textbf{Accumulated Damages on Perception:} Intuitively, the adverse weather conditions impair each agent's perception, so the multi-agent perception system might have accumulated damages~\cite{zhou2024v2xpnp} due to the agent-to-agent information sharing.   
\end{itemize} 

To solve the above three challenges, this paper proposes a novel Domain Generalization (DG) approach called \textit{V2X-DGW} for LiDAR-based 3D object detection of V2X multi-agent perception under adverse weather conditions. Our approach is designed to improve performance in unseen adverse weather conditions, training exclusively on the clean-weather source data. First of all, we design a new Adaptive Weather Augmentation (AWA) to mimic the LiDAR point clouds in adverse weather conditions with reduced perception range and diverse degradation. We then propose two alignment techniques for the generalizable representation learning, namely Trust-region Weather-invariant Alignment (TWA) and Agent-aware Contrastive Alignment (ACA). Specifically, TWA is proposed for domain generalization in the unseen adverse weather conditions and ACA is used as a contrastive learning based regularization to shrink the accumulated damages on perception due to adverse weather.

Furthermore, due to the difficulties of collecting multi-agent data with communication under adverse weathers, there are currently no publicized benchmark datasets to investigate this research problem. To facilitate this research, we leverage the publicized OPV2V~\cite{xu2022opv2v} and V2XSet~\cite{xu2022v2x} datasets of V2X based multi-agent perception, by simulating three common adverse weather conditions (Fog, Rain, Snow) by the physics-based point cloud models~\cite{hahner2021fog,kilic2021lidar,hahner2022lidar} to generate two new adverse weather V2X datasets, namely OPV2V-w and V2XSet-w. Finally, we conducted extensive experiments on our  simulated OPV2V-w and V2XSet-w datasets, to justify the effectiveness of our proposed method. Our contributions are summarized as follows.

\begin{itemize}
    \item To the best of our knowledge, we propose the \textbf{first research} of Domain Generalization for LiDAR-based V2X multi-agent perception systems under unseen adverse weather conditions. 
    
    \item To assess the impact of adverse weather on multi-agent perception systems, we simulate the effect of three common types of weather conditions(Fog, Rain, Snow), on point cloud from the clean  weather, utilizing the existing OPV2V and V2XSet datasets to create two new adverse weather benchmark datasets, namely OPV2V-w and V2XSet-w.

    \item We propose a new Adaptive Weather Augmentation (AWA) to mimic the unseen adverse weather conditions from the clean-weather source data only and design two new alignment methods for the generalizable representation learning: Trust-region Weather-invariant Alignment (TWA) for domain generalization in unseen adverse weather conditions and Agent-aware Contrastive Alignment (ACA) as a contrastive learning based regularization to relieve the accumulated damages by adverse weather. 
\end{itemize}

\section{Related Work}

\noindent \textbf{Multi-agent Perception.}
Multi-agent perception systems transcend the constraints of single-vehicle systems by utilizing information via collaboration modules, enhancing effectiveness and results of multi-agent perception tasks~\cite{xu2023bridging, xu2022opv2v, wang2020v2vnet, xu2022v2x,xu2022cobevt}.
Generally, three methods are mainly used for combining observations from multiple vehicles: merging raw data~\cite{chen2019cooper}, integrating features during processing~\cite{li2023s2r, chen2019f, wang2020v2vnet, xu2022opv2v, li2024advattack}, and output fusion~\cite{yu2022dair, arnold2020cooperative}.
Attfuse~\cite{xu2022opv2v} utilizes self-attention models to enhance the sharing features for V2V perception system.
SCOPE~\cite{yang2023spatio} presents a learning-based framework that targets multi-agent challenges, prioritizing the temporal context of the ego agent.
V2X-ViT~\cite{xu2022v2x,v2xvitv2} introduces a unified Vision Transformer architecture to integrate features from multi-agent perception systems.
While these methods demonstrate remarkable performance in multi-agent perception, they have primarily been evaluated under clear weather conditions, neglecting the impact of adverse weather.

\noindent \textbf{Domain 
Generalization.}~\label{sec:dg}
Domain generalization (DG) aims at achieving the excellent performance in unseen domains by exclusively learning from source domains. Current DG methods have been extensively explored on object detection~\cite{vidit2023clip, lehner20223d}, semantic segmentation~\cite{sanchez2023domain, li2023bev, kim2023single, xiao20233d,yan2024benchmarking} and reinforcement learning~\cite{hansen2021generalization, mazoure2022improving} tasks in 2D and 3D computer vision tasks.
As for strategies of DG employing multiple sources, which encompasses techniques such as the the disentanglement of domain-specific and domain-invariant features~\cite{liu2023d2ifln, zhang2022towards, bui2021exploiting}, alignment of source domain distributions~\cite{chen2023domain, matsuura2020domain, li2018domain}, and domain augmentation~\cite{zhou2020learning}. 
These approaches have been instrumental in navigating the complexities associated with learning from diverse source domains. Some recent studies~\cite{zhou2022domain, xiao20233d, kim2023single} have analyzed the sparsity of point clouds in addressing DG issues. However, the extension of DG to the LiDAR-based 3D object detection on multi-agent perception systems in adverse weather conditions remains relatively untapped. This paper aims to design a new DG method to promote the LiDAR-based 3D object detection on multi-agent perception systems in adverse weather conditions.

\section{Methodology}

\begin{figure*}[!t]
\centering
\subfloat{ 
\includegraphics[width=1\textwidth]{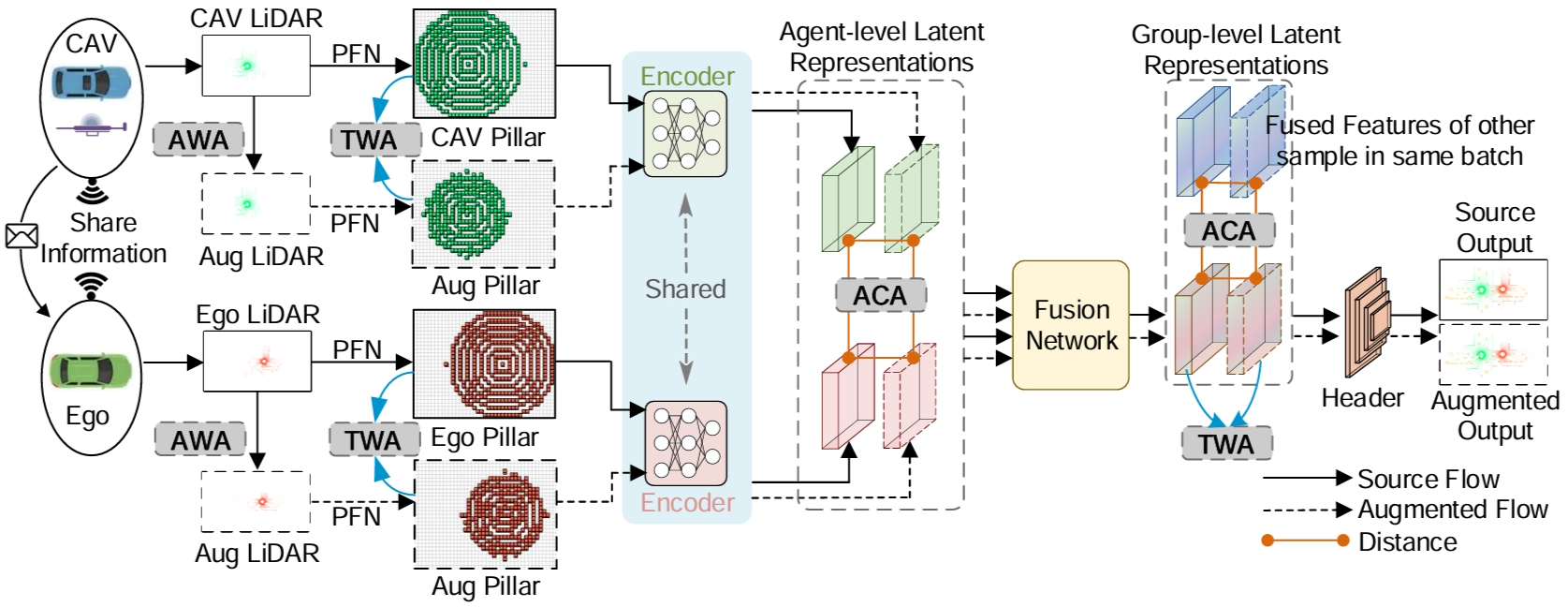}%
}
\caption{\textbf{Overview framework of the proposed V2X-DGW for domain generalization of multi-agent perception system in adverse weather.} During the training stage, every agent has source flow (clean weather) and augmented flow (mimicked adverse weather). Pillar features are obtained from the Pillar Feature Net (PFN)~\cite{lang2019pointpillars}, which is shown in bird view here.}
\label{fig:method}
\vspace{-1.5em}
\end{figure*}

\subsection{Adverse-Weather Benchmark for Multi-agent Perception}\label{subsec:weather}
Existing multi-agent perception datasets are all collected under the assumption of clean weather. We employ state-of-the-art physics-based simulation methods on the existing multi-agent perception datasets (OPV2V~\cite{xu2022opv2v} and V2Xset~\cite{xu2022v2x}) to create two new benchmarks: OPV2V-w and V2Xset-w, to assess the impact of harsh weather.

\noindent \textbf{Physics-based Simulation.}
We consider three different harsh weather conditions: \textit{Fog}, \textit{Rain}, \textit{Snow}, as they are common in the real world. Physics-based point cloud simulation methods have shown its success to approximate real weather conditions. We adopt three state-of-the-art physics-based point cloud simulation methods~\cite{hahner2021fog,kilic2021lidar,hahner2022lidar} for simulating \textit{Fog}\footnote{\href{https://github.com/MartinHahner/LiDAR\_fog\_sim}{https://github.com/MartinHahner/LiDAR\_fog\_sim}}, \textit{Rain}\footnote{\href{https://github.com/velatkilic/LISA}{https://github.com/velatkilic/LISA}} and \textit{Snow}\footnote{\href{https://github.com/SysCV/LiDAR\_snow\_sim}{https://github.com/SysCV/LiDAR\_snow\_sim}} conditions respectively with the publicized source code. These simulation methods are based on the physical reflection and geometrical optical model in  LiDAR sensors. The most recent study~\cite{dong2023benchmarking} have corroborated the consistency of perception performance in synthetic weather data  by the physics-based simulation with that observed in real-world  data under adverse weather scenarios.

\noindent \textbf{OPV2V-w and V2XSet-w.}
OPV2V~\cite{xu2022opv2v} and V2XSet~\cite{xu2022v2x} are two large scale datasets for LiDAR based multi-agent  perception. Since both of them are collected from the same game engine software Carla~\cite{dosovitskiy2017carla}, leveraging these datasets to  focus on the impact of weather variations later. We deploy physics-based simulation methods on the testing sets of OPV2V and V2XSet, then obtain OPV2V-w and V2XSet-w to evaluate domain generalization under adverse weather. 

\subsection{Overview of Architecture}
\begin{figure}[!t]
\centering
\subfloat{ 
\includegraphics[width=1\columnwidth]{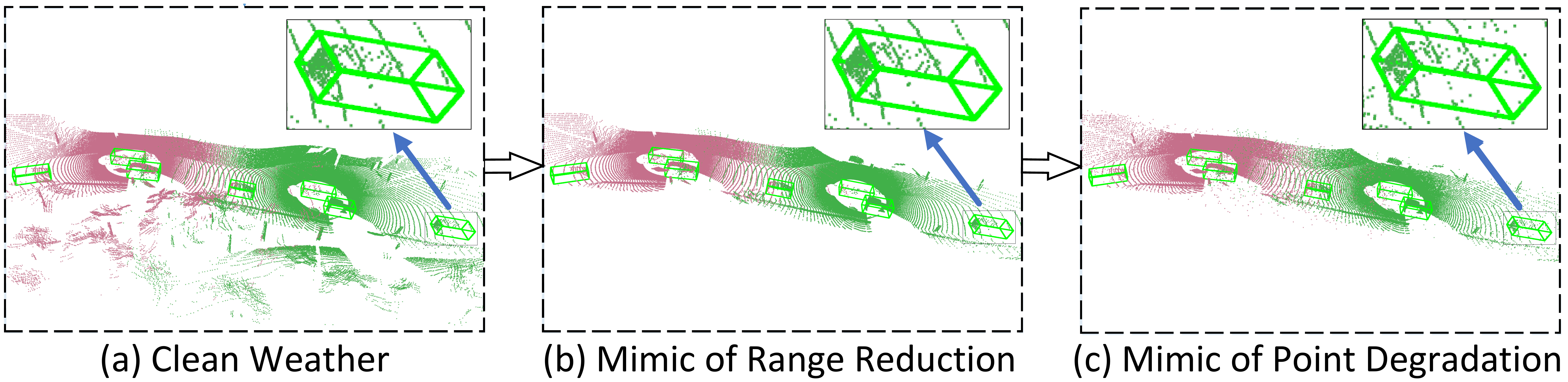}%
}
\caption{\textbf{Illustration of simulation before and after Adaptive Weather Augmentation.}}
\label{fig:method_awa}
\vspace{-1.5em}
\end{figure}

\begin{figure*}[!t]
\centering
\subfloat{ 
\includegraphics[width=1.65\columnwidth]{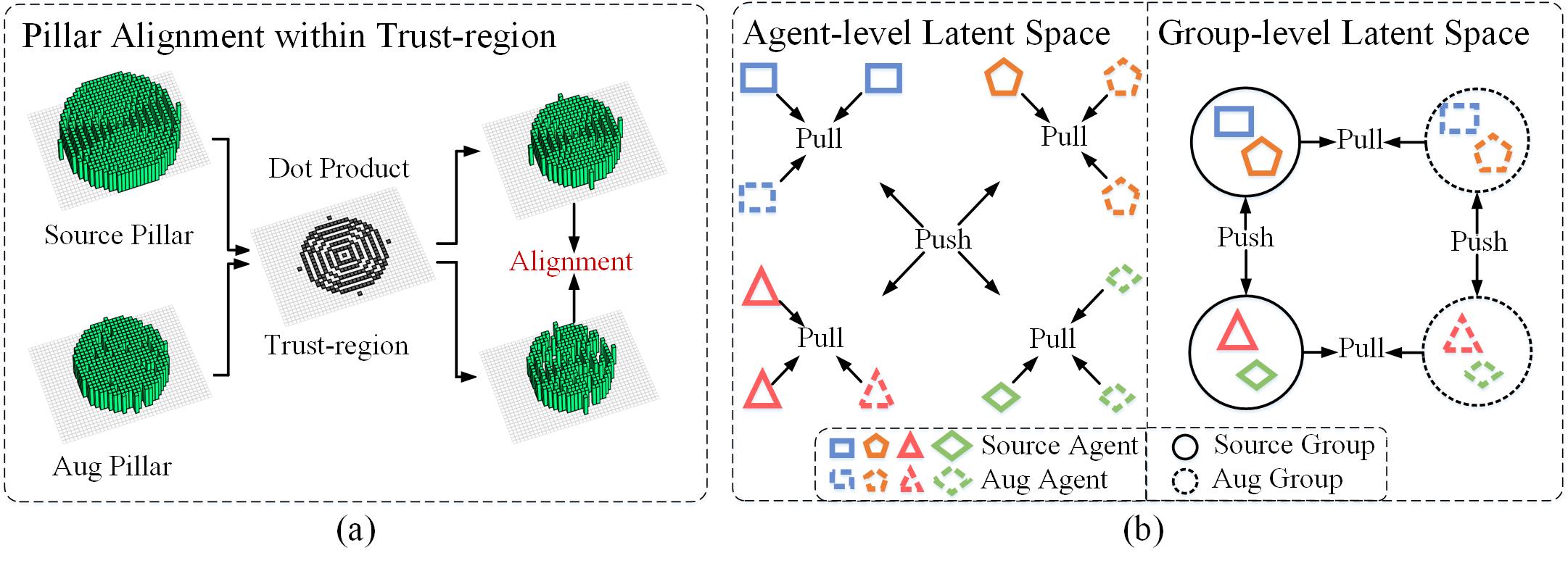}%
}
\vspace{-1em}
\caption{\textbf{Illustration of TWA and ACA.} (a) Pillar-based weather-invariant alignment within Trust-region, (b) Agent-aware contrastive alignment in agent-level and group-level. Best view in color.}
\label{fig:method_twa_aca}
\vspace{-1.5em}
\end{figure*}

Our V2X-DGW is depicted in Fig.~\ref{fig:method}. Given a spatial graph of CAVs within the communication range in clean weather source data, the point clouds of ego and CAVs are denoted $\mathbf{P}^{s}_{ego} \in \mathbb{R}^{4 \times m}$ and $\mathbf{P}^{s}_{cav} \in \mathbb{R}^{4 \times m}$, respectively. We firstly deploy Adaptive Weather Augmentation $\mathbf{A}_{w}$ on the clean weather data and obtain the augmented data $\mathbf{P}^{a}_{ego} \in \mathbb{R}^{4 \times m}$ and $\mathbf{P}^{a}_{cav} \in \mathbb{R}^{4 \times m}$. 
Then all of them are projected into 2D pseudo-image shape $\mathbf{I} \in \mathbb{R}^{C \times H \times W}$ by leveraging the PointPillar~\cite{lang2019pointpillars}'s Pillar Feature Net (PFN), defined as $\mathbf{M}$. So, $\mathbf{I}^{s}_{cav},\mathbf{I}^{a}_{cav},\mathbf{I}^{s}_{ego},\mathbf{I}^{a}_{ego} = \mathbf{M}(\mathbf{P}^{s}_{cav},\mathbf{A}_{w}(\mathbf{P}^{s}_{cav}), \mathbf{P}^{s}_{ego},  \mathbf{A}_{w}(\mathbf{P}^{s}_{ego})) 
\label{eq:pointpillar}.$ In our framework, each CAV shares the same Encoder for LiDAR feature extraction in all conditions. The ego vehicle acquires the visual features of neighboring CAVs through wireless communication. The intermediate features aggregated from $N$ surrounding CAVs in source and augmented data are denoted as $\mathbf{F}^{s}_{cav} \in \mathbb{R}^{N \times H \times W \times C}$ and $\mathbf{F}^{a}_{cav} \in \mathbb{R}^{N \times H \times W \times C}$, respectively. The ego intermediate features are denoted as $\mathbf{F}^{s}_{ego} \in \mathbb{R}^{1 \times H \times W \times C}$ and $\mathbf{F}^{a}_{ego} \in \mathbb{R}^{1 \times H \times W \times C}$, respectively. The intermediate features are then fused by the same Feature Fusion Network (FFN). Finally, the fused feature maps are fed into the same prediction header for 3D bounding-box regression and classification. Here we formulate the whole framework $\mathbf{DGW}(\cdot)$ as 
\begin{equation}    \mathbf{F}^{s}_{cav},\mathbf{F}^{a}_{cav},\mathbf{F}^{s}_{ego},\mathbf{F}^{a}_{ego} = \mathbf{E}(\mathbf{T}_{w}(\mathbf{I}^{s}_{cav},\mathbf{I}^{a}_{cav}), \mathbf{T}_{w}(\mathbf{I}^{s}_{ego},\mathbf{I}^{a}_{ego})), 
    \label{eq:encoder}
\end{equation}
\begin{equation}
    \hat{\mathbf{F}}^{s}, \hat{\mathbf{F}}^{a} = \text{FFN}(\mathbf{C}_{a}(\mathbf{F}^{s}_{cav},\mathbf{F}^{a}_{cav},\mathbf{F}^{s}_{ego},\mathbf{F}^{a}_{ego})), 
    \label{fusion network}
\end{equation}
\begin{equation}
    \mathbf{DGW} (\mathbf{P}^{s}_{cav},\mathbf{P}^{s}_{ego}) =  \mathbf{H}(\mathbf{C}_{g}(\mathbf{T}_{w}(\hat{\mathbf{F}}^{s}, \hat{\mathbf{F}}^{a}),*)),
\end{equation}
where $\mathbf{E}(\cdot)$ is the Encoder for LiDAR feature extraction, $\text{FFN}(\cdot)$ is the Feature Fusion Network responsible for fusing the features of CAVs and the ego vehicle, $\hat{\mathbf{F}}$ is the fused feature extracted from $\text{FFN}(\cdot)$ and $\mathbf{H}$ is the prediction header for 3D object detection, $\mathbf{T}_{w}(\cdot)$ is the proposed Trust-region Weather-invariant Alignment (TWA), $\mathbf{C}_{a}(\cdot)$ is the proposed Agent-aware Contrastive  Alignment (ACA) in the agent-level,  $\mathbf{C}_{g}(\cdot)$ is the proposed Agent-aware Contrastive  Alignment (ACA) in the group-level, and $*$ indicates the fused features of other samples in the same batch.

\subsection{Adaptive Weather Augmentation}
Based on our above discovery, the adverse weather conditions have 1) perception range reduction and 2) point cloud degradation in the LiDAR based point cloud data. We design Adaptive Weather Augmentation (AWA) from two views.

\noindent \textbf{Mimic of Perception Range Reduction.} Given $\mathbf{P}^{s} \in \mathbb{R}^{4 \times m}$ that contains a set of 3D points \{$P_i = (x_{i}, y_{i}, z_{i})  |  i = 1,2,3,...,m$\}. The reception of LiDAR signals is subject to certain range limitations, thereby imposing maximal values on the x, y, and z coordinates of the point cloud data, denoted as $x_{m}, y_{m}, z_{m}$, respectively. Then, we sample the  point clouds to reduce perception range by 

\begin{equation}
\begin{split}
    \mathbf{P}^{r} & = \mathbf{S}_{x,y,z} (\mathbf{P}^{s}) \\
     \text{s.t.} & \quad |x_{i}/x_{m}| \leq \delta_{x} , |y_{i}/y_{m}| \leq \delta_{y}, |z_{i}/z_{m}| \leq \delta_{z},
\end{split}
\end{equation}
where $\mathbf{S}_{x,y,z}(\cdot)$ is the sampling function, $\delta_{x}, \delta_{y}, \delta_{z}$ are the thresholds of LiDAR receiving range and $\delta_{x}, \delta_{y}, \delta_{z} \sim U(\phi_{L},\phi_{U})$. $\phi_{L}$ and $\phi_{U}$ are the lower bound and upper bound of random sampling, respectively.

\noindent \textbf{Mimic of Point Cloud Degradation.}  After the reception range decrease, we deploy three random degradation functions: random dropout $\Gamma (\cdot)$, random jittering $\Delta (\cdot)$ and random noise perturbation $\Theta (\cdot)$. The final AWA output  $\mathbf{P}^{a}$ can be obtained as 
\begin{equation}
    \mathbf{P}^{a} = \Theta(\Delta(\Gamma(\mathbf{P}^{r}))). 
\end{equation}

The AWA process is illustrated in Fig.~\ref{fig:method_awa}, which is  dynamically executed in real-time during each training iteration. This procedure generates augmented point cloud data, which is exclusively utilized for the purpose of simulating the unseen adverse weather conditions, so as to improve the model generalization capability.

\subsection{Trust-region Weather-invariant Alignment (TWA)}
To improve the generalization of multi-agent perception systems in adverse weather, aligning augmented flow (mimicked adverse weather) features with source flow (clean weather) features is essential. However, the direct alignment on the whole features might risk model training. Thus, we propose a trust-region weather-invariant alignment to guide the model focus on the reduced perception range.   

\noindent \textbf{Trust-region Weight.} Most of current multi-agent perception systems leverage PointPillar~\cite{lang2019pointpillars} to process point clouds into pseudo-image, which is also called pillar and named as $\mathbf{I} \in \mathbb{R}^{C \times H \times W}$ and fed into 
Convolutional Neural Networks based Encoder. To address variations in point clouds from augmented and source settings, we define a trust-region  $\mathbf{T} \in \mathbb{R}^{H \times W}$ on the pillar for weather-invariant learning, which is a 2D matrix that determines whether a perceived location should be considered or not. The each element of $\mathbf{T}$ could be calculated by
\begin{multline}
\mathbf{T}(h, w) = \begin{cases} 
1 & \text{if } \exists c \in \{1, \ldots, C\} : \mathbf{I}^{s}(c,h,w) \neq 0 \land \\
  & \mathbf{I}^{r}(c,h,w) \neq 0 \\  
0 & \text{otherwise}. 
\end{cases}
\end{multline}

\noindent \textbf{Pillar Alignment within Trust-region.}
We align each agent's pillar features between source and augmented data, ensuring the Encoder to extract deep features within the trust region as shown in Fig.~\ref{fig:method_twa_aca}(a):
\begin{equation}
\mathcal{L}_{PAT} = \sum_{h=1}^{H} \sum_{w=1}^{W} \mathbf{T}(h, w) \cdot \sum_{c=1}^{C} || \mathbf{I}^s(c, h, w) - \mathbf{I}^a(c, h, w) ||_1.
\end{equation}

\noindent \textbf{Fused Feature Alignment.}
After Feature Fusion Network (FFN), we align the augmented fused  features with source fused features. This constraint enhances system robustness by minimizing weather-related disparities, defined as
\begin{equation}
\mathcal{L}_{FFA} = \sum_{h=1}^{H} \sum_{w=1}^{W} \sum_{c=1}^{C} ||  \hat{\mathbf{F}}^{s}(c,h,w)-  \hat{\mathbf{F}}^{a}(c,h,w)||_1.
\end{equation}
The total loss of TWA is formulated as 
   $\mathcal{L}_{TWA} = \alpha_1 \mathcal{L}_{PAT} + \alpha_2 \mathcal{L}_{FFA}$, where $\alpha_1$ and $\alpha_2$ are balance coefficients. 

\subsection{Agent-aware Contrastive Alignment (ACA)}

We propose Agent-aware Contrastive Alignment (ACA) as contrastive learning~\cite{khosla2020supervised} based regularization to enhance the model robustness so as to relieve the accumulated damages on  multi-agent perception under the adverse weather conditions.

\noindent \textbf{ACA in Agent-level.} Within a given scene, each agent/CAV is unique with a distinct ID. As shown in Fig.~\ref{fig:method_twa_aca}(b), we pull the feature distance of the same agent ID of the source and augmented data in the latent space, and we push the feature distance of the different agent IDs of the source and augmented data in the latent space. The contrastive loss of ACA in agent-level is formulated as

\begin{multline}
\mathcal{L}_{ACA_a} = -\frac{1}{B_a} \sum_{i=1}^{B_a} \sum_{p \in P(i)} \Bigg\{ 
\log \Big[ \exp\left(\mathbf{F}_i^s \cdot \mathbf{F}_p^s \slash {\tau}\right) \\
+ \exp\left(\mathbf{F}_i^s \cdot \mathbf{F}_p^{a}\slash{\tau}\right) 
+ \exp\left(\mathbf{F}_i^a \cdot \mathbf{F}_p^a\slash{\tau}\right) \Big] \\
- \log \Big[ \sum_{j=1}^{B_a} \exp\left(\mathbf{F}_i^s \cdot \mathbf{F}_j^s\slash{\tau}\right) 
+ \sum_{k=1}^{B_a} \exp\left(\mathbf{F}_i^a \cdot \mathbf{F}_k^a\slash{\tau}\right) \Big] 
\Bigg\},
\end{multline}
where $B_a$ is the agent-level batch size in the training, $P(i)$ is the set of positive samples (the same-ID agent), $\tau$ is the temperature hyperparameter.
 
\noindent \textbf{ACA in Group-level.} 
After the FFN based fusion, the fused feature  of multiple agents, $\hat{\mathbf{F}}^s$ or  $\hat{\mathbf{F}}^a$, can be understood as group-level unit. As shown in Fig.~\ref{fig:method_twa_aca}(b), we pull the feature distance of the same group between the source and augmented data in the latent space, and we push the feature distance of the different groups within the source or augmented data in the latent space. The contrastive loss of ACA in group-level is  formulated as

\begin{multline}
\mathcal{L}_{ACA_g} = -\frac{1}{B_g} \sum_{i=1}^{B_g} \Biggl\{ 
\log \Biggl[ \exp\left(\hat{\mathbf{F}}_i^s \cdot \hat{\mathbf{F}}_i^a \slash {\tau}\right) \Biggr] \\
- \log \Biggl[ 
\sum_{j=1}^{B_g} \exp\left(\hat{\mathbf{F}}_i^s \cdot \hat{\mathbf{F}}_j^s \slash {\tau}\right) 
+ \sum_{k=1}^{B_g} \exp\left(\hat{\mathbf{F}}_i^a \cdot \hat{\mathbf{F}}_k^a \slash {\tau}\right) 
\Biggr]
\Biggr\},
\end{multline}
where $B_g$ is group-level batch size in training. The ACA loss function can be formulated as 
$\mathcal{L}_{ACA} = \beta_1 \mathcal{L}_{ACA_a} + \beta_2 \mathcal{L}_{ACA_g}$, where $\beta_1$ and $\beta_2$ are balance weights. 

\begin{table*}[htb]
\caption{\textbf{3D detection performance from clean weather to adverse weather.} The training process utilizes only the source data (clean weather). During inference, the trained model can be applied to clean, foggy, rainy, and snowy weathers.}
\resizebox{1\textwidth}{!}{%
\begin{tabular}{@{}c|cc|cccccc|cccccc@{}}
\toprule
\textbf{}                    & \multicolumn{2}{c|}{\textbf{OPV2V}} & \multicolumn{6}{c|}{\textbf{OPV2V-w}}                                          & \multicolumn{6}{c}{\textbf{V2XSet-w}}                                         \\ \midrule
                             & \multicolumn{2}{c|}{clean}          & \multicolumn{2}{c}{fog} & \multicolumn{2}{c}{rain} & \multicolumn{2}{c|}{snow} & \multicolumn{2}{c}{fog} & \multicolumn{2}{c}{rain} & \multicolumn{2}{c}{snow} \\
\multirow{-2}{*}{Method}     & AP@0.5           & AP@0.7           & AP@0.5     & AP@0.7     & AP@0.5      & AP@0.7     & AP@0.5      & AP@0.7      & AP@0.5     & AP@0.7     & AP@0.5      & AP@0.7     & AP@0.5      & AP@0.7     \\ \midrule
Baseline                         & 88.10            & 79.48            & 67.55      & 59.40      & 71.17       & 58.34      & 60.51       & 46.96       & 63.04      & 55.92      & 65.05       & 53.64      & 63.86       & 52.47      \\
IBN-Net~\cite{pan2018two}                         & 85.08            & 77.07            & 66.45      & 58.31      & 68.92       & 59.55      & 55.28       & 45.80       & 61.20      & 52.92      & 62.27       & 53.54      & 61.03       & 51.94      \\
MLDG$_a$~\cite{li2018learning}                      & 87.68            & 78.13            & 67.23      & 59.30      & 68.55       & 57.50      & 58.86       & 47.67       & 63.12      & 55.07      & 63.15       & 53.38      & 62.06       & 52.29      \\
MLDG$_b$~\cite{li2018learning}                      & 83.93            & 73.49            & 62.73      & 54.52      & 63.94       & 54.39      & 51.52       & 42.78       & 58.36      & 51.01      & 57.17       & 48.88      & 55.88       & 47.55      \\
Ours & \textbf{89.23}            & \textbf{81.52}            & \textbf{71.35}      & \textbf{64.35}      & \textbf{75.31}       & \textbf{67.98}      & \textbf{65.80}       & \textbf{58.38}       & \textbf{66.66}      & \textbf{60.37}      & \textbf{70.61}       & \textbf{63.77}      & \textbf{69.44}       & \textbf{62.68}      \\ \midrule
GRL~\cite{ganin2015unsupervised}                          & 87.47            & 79.23            & 66.40      & 59.07      & 70.38       & 58.24      & 57.32       & 48.32       & 61.81      & 55.21      & 64.30       & 53.90      & 63.00       & 53.19      \\
AdvGRL~\cite{li2023domain}                       & 88.20            & 80.62            & 68.11      & 60.87      & 71.76       & 62.92      & 60.76       & 51.72       & 64.14      & 57.80      & 67.33       & 59.42      & 66.12       & 57.74  \\
S2R-AFA~\cite{li2023s2r}                      & 87.34            & 78.65            & 67.53      & 59.30      & 70.64       & 57.62      & 60.43       & 46.68       & 63.04      & 55.75      & 64.75       & 53.06      & 63.54       & 51.90      \\ \bottomrule \end{tabular}}
\label{tab:o2all}
\end{table*}

\subsection{Overall Loss Function}
Following~\cite{li2023s2r}, we use the focal loss~\cite{lin2017focal} and smooth $L_1$ loss as the detection loss. Our final loss is the combination of detection loss and alignment losses as  $\mathcal{L}_{total} = \mathcal{L}^{s}_{det} + \mathcal{L}^{a}_{det} + \mathcal{L}_{TWA} + \mathcal{L}_{ACA}$.

\section{Experiments}

\subsection{Experiment Setup}

\noindent \textbf{Dataset.} We utilize two popular multi-agent perception datasets.
OPV2V~\cite{xu2022opv2v} is a comprehensive simulated dataset designed for V2V cooperative perception tasks, collected through the integration of the CARLA~\cite{dosovitskiy2017carla} and OpenCDA~\cite{10045043} platforms. It encompasses 73 varied scenes with CAVs. The dataset includes 6,764 frames for training and 2,719 frames for testing. V2XSet~\cite{xu2022v2x} is an expansive simulated dataset for V2X cooperative perception task, is also assembled utilizing CARLA. The dataset is organized into training and testing subsets, containing 6,694 and 2,833 frames  respectively. We simulate adverse weather conditions only on their  \textit{testing sets} to generate OPV2V-w and V2XSet-w datasets for our experiments, as detailed in Section~\ref{subsec:weather}.

\noindent \textbf{Evaluation Metrics.}
The final 3D vehicle detection accuracy are selected as our performance evaluation. Following~\cite{xu2022opv2v,xu2022v2x}, we set the evaluation range as $x\in[-140, 140]$ meters,  $y\in[-40, 40]$ meters, where all CAVs are included in this spatial range of the experiment. We measure the accuracy with Average Precisions (AP) at Intersection-over-Union (IoU) threshold of $0.5$ and $0.7$.

\noindent \textbf{Compared Methods.}
To analyze the effect of adverse weather conditions on the perception performance of multi-agent system, two Domain Generalization (DG) methods, \ie, IBN-Net~\cite{pan2018two}, and  MLDG~\cite{li2018learning}, and three Domain Adaptation (DA) methods, \ie, GRL~\cite{ganin2015unsupervised}, AdvGRL~\cite{li2023domain}, and S2R-AFA~\cite{li2023s2r} are implemented to be evaluated. For the MLDG~\cite{li2018learning} setting, we employ two different strategies: randomly splitting each batch into meta-train and meta-test sets, denoted as MLDG$_a$, and splitting data by maximizing feature distances, denoted as MLDG$_b$. As for the domain adaptation methods, we follow the settings in~\cite{li2023s2r}.

\noindent \textbf{Experimental Settings.} We evaluate the detection performance across four weather conditions: clean, fog, rain and snow. All methods in the experiments employ the same cooperative perception method AttFuse~\cite{xu2022opv2v} (as FFN), with the PointPillar based Encoder for fair comparison. We denote the AttFuse method without DA or DG  as our Baseline, which is only trained on source data (clean weather). For implementation of our V2X-DGW method, $\phi_{L}$ and $\phi_{U}$ are set to 0.5 and 0.8, $\alpha_1$ and $\alpha_2$ are set to 0.1 and 1, $\beta_1$ and $\beta_2$ are set to 0.01 and 0.01, $\tau$ is set to 0.07. All DG methods are trained on the OPV2V training set (clean weather) only. For DA methods, the labeled OPV2V training set serves as the source domain, while the unlabeled OPV2V training set added with fog by physics-based simulation is utilized as the target domain during training. These methods are all tested/evaluated on the OPV2V-w and V2XSet-w with fog, rain, snow weather conditions.

\subsection{Quantitative Evaluation}
\noindent \textbf{3D Detection Performance.}
Table~\ref{tab:o2all} shows the experiment performance. Without the effect of the adverse weathers, the Baseline without any DA or DG methods can gain a outstanding performance on OPV2V clean testing set. However, when it is deployed on OPV2V-w testing sets, it drops $27.59\%$ / $32.52\%$ for AP@0.5/0.7 on the snow weather of OPV2V-w. It indicates that the cooperative perception system has significantly detrimental effects on performance under adverse weather conditions, which also underscores the importance of generalization capabilities.
After applying the DA and DG methods, only a few have shown improved performance on the unseen weather domains. This could be attributed to the differences in domain gaps between 2D images and point clouds, as well as the ambiguity of appearance features in the LiDAR domain. Consequently, the performance enhancement achieved through learning appearance-invariant features may not be sufficient. While our method consistently outperforms comparative benchmarks in both clean and adverse weather conditions. It underscores the generalization capabilities of our proposed method for cooperative perception across a variety of adverse weather conditions.

\begin{figure*}[!t]
\centering
\subfloat{ 
\includegraphics[width=1 \textwidth]{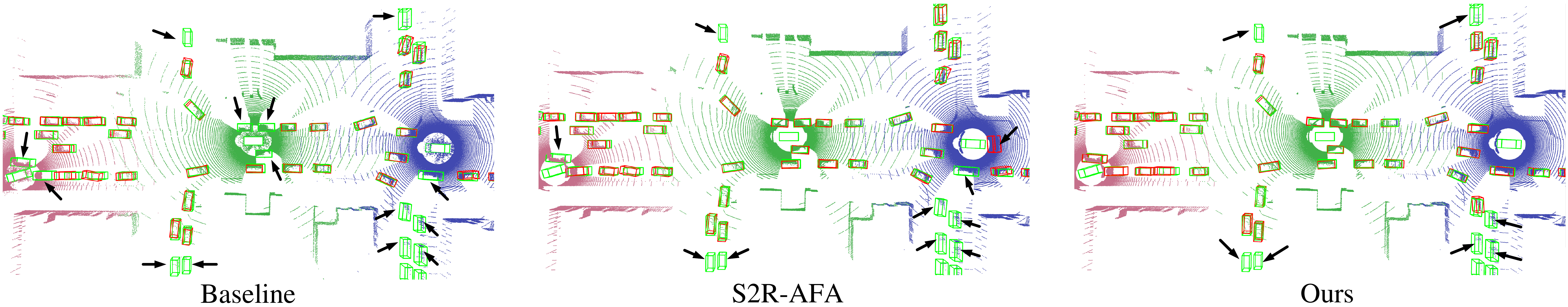}%
} 
\caption{\textbf{3D object detection visualization example  under adverse weather.} \textcolor{green}{Green} and \textcolor{red}{red} 3D bounding boxes represent the \textcolor{green}{ground truth} and \textcolor{red}{prediction} respectively. The detection errors are highlighted using black  arrows. OPV2V-w fog weather is used as example here.}
\label{fig:det_vis}
\vspace{-1em}
\end{figure*}

\noindent \textbf{3D Detection Visualization.}
The Fig.~\ref{fig:det_vis} presents the 3D detection visualization on OPV2V-w fog subset of the Baseline, S2R-AFA~\cite{li2023s2r}, and our proposed method. When generalizing a cooperative perception model trained in clear weather to adverse weather conditions, both the baseline and DA methods exhibit detection errors, highlighting the detrimental impact of harsh weather on cooperative perception. More critically, our method succeeds in identifying a greater number of dynamic objects, as evidenced by a higher match rate with ground-truth. This underscores the superior generalization capabilities of our proposed method.

\begin{figure}[!t]
\centering
\subfloat{ 
\includegraphics[width=1\columnwidth]{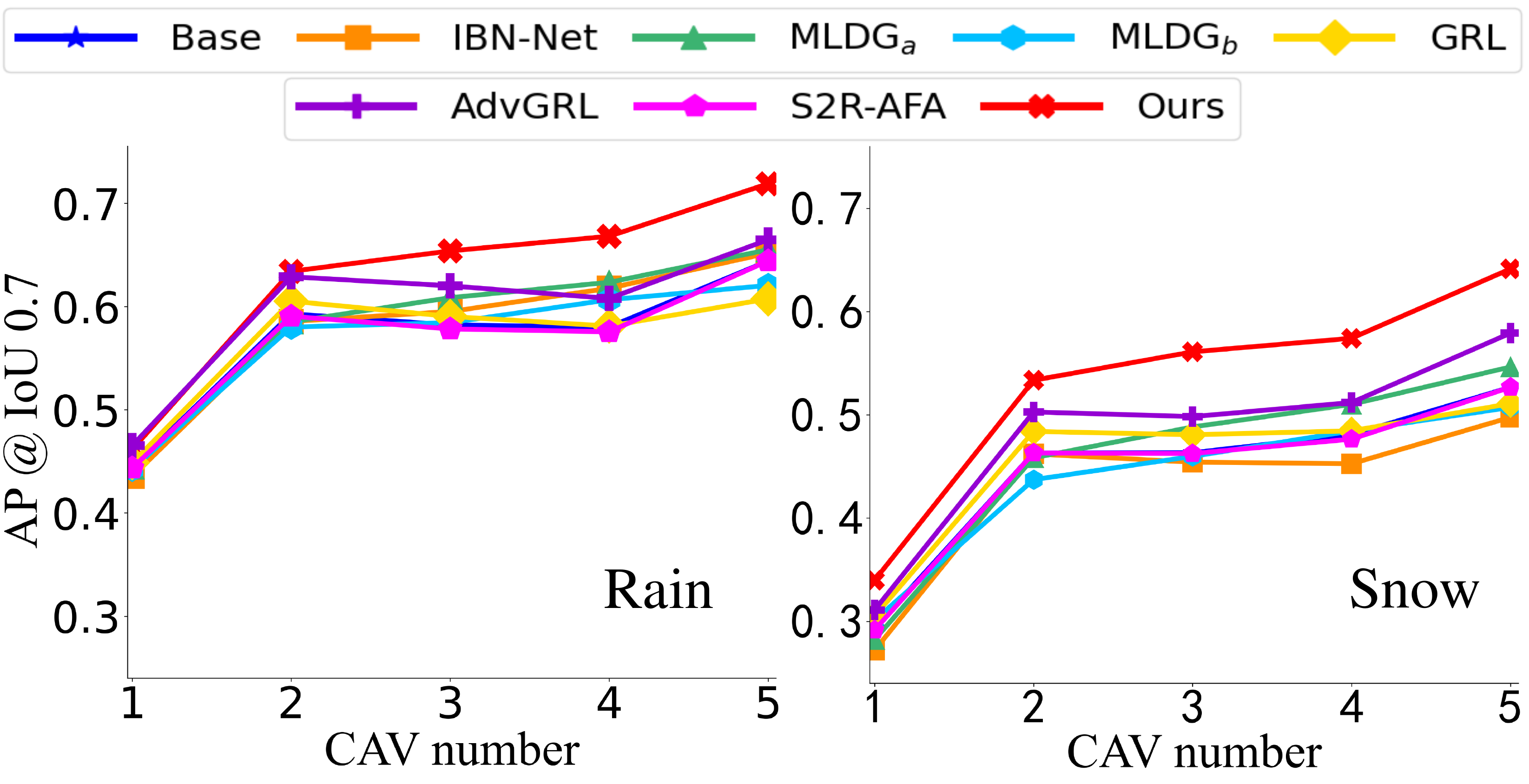}%
}
\caption{\textbf{Sensitivity to agent numbers.} Other methods' decreasing detection performance with more agent numbers on OPV2V-w (contrary to intuition~\cite{xu2022opv2v}) indicates the accumulated damages under adverse weather, while our method is more robust.}
\vspace{-1.5em}
\label{fig:cavnum}
\end{figure}

\noindent \textbf{Sensitivity to Agent Numbers.} The Fig.~\ref{fig:cavnum} illustrates the impact of cooperative perception performance with varying numbers of agents under rain and snow weather. As shown in the intuition~\cite{xu2022opv2v}, the cooperative perception performance should be improved with an increasing number of nearby agents. However, most comparison methods sometimes have decreasing cooperative detection performance with increasing agent numbers under adverse weather scenarios, which is because of their sensitivity to the accumulated damages from more agents in adverse weather. Differently, our V2X-DGW could overcome the challenge of accumulated damages.

\noindent \textbf{Ablation Studies.}
To validate the effectiveness of our proposed components, we provide the quantitative comparison on detection result under clean, fog, rain, snow weather conditions in Table~\ref{tab:ablation}. Each component contributes to the model's performance enhancement across all four weather conditions, validating the rationale behind the design of AWA, TWA, and ACA.

\begin{table}[]
\centering
\caption{\textbf{Ablation study.} Effects of the AWA, TWA, ACA components. The first row is the result by  Baseline.}
\label{tab:ablation}
\resizebox{\columnwidth}{!}{%
\begin{tabular}{@{}ccc|cccccccc@{}}
\toprule
\multirow{2}{*}{AWA} & \multirow{2}{*}{TWA} & \multirow{2}{*}{ACA} & {clean} & {fog} & {rain} & {snow} \\
                     &                       &                      & AP@0.5/0.7      & AP@0.5/0.7     & AP@0.5/0.7     & AP@0.5/0.7     \\ \midrule
                     &                       &                      & 88.10/79.48       & 67.54/59.39      & 71.17/58.33      & 60.51/46.96      \\
\midrule                     
\checkmark                    &                       &                      & 87.03/79.31       & 69.42/62.58      & 74.16/66.50      & 64.14/56.73      \\
\checkmark                    & \checkmark                     &                      & 88.88/\textbf{81.15}       & 70.83/63.71      & 74.36/67.10      & 64.65/57.21      \\
\checkmark                    &                       & \checkmark                    & 88.66/80.80       & 70.95/63.74      & 74.90/67.42      & 65.06/57.63      \\
\checkmark                    & \checkmark                     & \checkmark                    & \textbf{89.23}/\textbf{81.15}       & \textbf{71.35}/\textbf{64.34}      & \textbf{75.30}/\textbf{67.97}      & \textbf{65.80}/\textbf{58.38}      \\ \bottomrule
\end{tabular}}
\vspace{-0.5em}
\end{table}

\section{Conclusions}\label{Sec:Conclusions}

In this paper, we propose a domain generalization approach for LiDAR-based 3D object detection on the multi-agent perception systems under the adverse weather conditions (Fog, Rain, Snow). To investigate the  weather impact on multi-agent perception, the effect of three common adverse weather conditions are simulated on the two publicized V2X datasets, and to obtain two new V2X based adverse-weather multi-agent perception  datasets: OPV2V-w and V2XSet-w. To address the range reduction, point cloud degradation and accumulated damages in adverse weather, we first introduce the Adaptive Weather Augmentation (AWA) to simulate the unseen adverse weather domain, and then propose two alignments for generalizable representation learning: Trust-region  Weather-invariant Alignment (TWA) and Agent-aware Contrastive Alignment (ACA). Extensive experiments show that our method achieved outstanding 3D detection performance in the unseen adverse weather conditions.

\bibliographystyle{IEEEtran}
\bibliography{baolu}

\end{document}